\documentclass{article}
\usepackage{arxiv}

\usepackage{times}
\usepackage{soul}
\usepackage{url}
\usepackage[hidelinks]{hyperref}
\usepackage[utf8]{inputenc}
\usepackage[small]{caption}
\usepackage{graphicx}
\usepackage{amsmath}
\usepackage{amsthm}
\usepackage{booktabs}
\usepackage{algorithm}
\usepackage{algorithmic}
\usepackage[switch]{lineno}

\usepackage{amsmath}
\usepackage{amssymb}
\usepackage{xspace}
\usepackage{array}

\newcommand{\pil}{PIL } 
\newcommand{\dl}{DL } 
\newcommand{\il}{IL } 
\newcolumntype{R}{>{\raggedleft\arraybackslash}p{1.3cm}}
\newcolumntype{B}{>{\raggedleft\arraybackslash}p{1.3cm}}
\newcommand{\ade}[2][n]{%
    \ifx n#1\xspace{}minADE\(_{#2}\)\xspace \else
    \ifx b#1\xspace{}\textbf{minADE\(\mathbf{_{#2}}\)}\xspace 
    \fi\fi
    }
\newcommand{\fde}[2][n]{%
    \ifx n#1\xspace{}minFDE\(_{#2}\)\xspace  \else
    \ifx b#1\xspace\textbf{minFDE\(\mathbf{_{#2}}\)}\xspace 
    \fi\fi
    }
\newcommand{\hrate}[3][n]{%
    \ifx n#1\xspace{}HitRate\(_{#2,#3}\)\xspace \else
    \ifx b#1\xspace{}\textbf{HitRate\(\mathbf{_{#2,#3}}\)}\xspace
    \fi\fi
    }

\newcommand{\tpm}{$\pm$}

\title{Informed Spectral Normalized Gaussian Processes for Trajectory Prediction}

\author{\hspace{1mm}Christian Schlauch \\
	Humboldt-Universit\"at zu Berlin, \\ 
	and Continental AG\\
	Berlin, Germany\\ 
	\And 
	\hspace{1mm}Christian Wirth\\
	Continental AG\\
	Frankfurt am Main, Germany
	\And
	\hspace{1mm}Nadja Klein \\
	Technische Universit\"at Dortmund \\ Chair of Uncertainty Quantification and Statistical Learning\\
	Berlin, Germany
}

\begin{document}

\maketitle

\begin{abstract}
Prior parameter distributions provide an elegant way to represent prior expert and world knowledge for informed learning. Previous work has shown that using such informative priors to regularize probabilistic deep learning (DL) models increases their performance and data-efficiency. However, commonly used sampling-based approximations for probabilistic DL models can be computationally expensive, requiring multiple inference passes and longer training times. Promising alternatives are compute-efficient last layer kernel approximations like spectral normalized Gaussian processes (SNGPs). We propose a novel regularization-based continual learning method for SNGPs, which enables the use of informative priors that represent prior knowledge learned from previous tasks. Our proposal builds upon well-established methods and requires no rehearsal memory or parameter expansion. We apply our \textit{informed SNGP} model to the trajectory prediction problem in autonomous driving by integrating prior drivability knowledge. On two public datasets, we investigate its performance under diminishing training data and across locations, and thereby demonstrate an increase in data-efficiency and robustness to location-transfers over non-informed and informed baselines.

\end{abstract}

\section{Introduction}
Deep learning (DL) has become a powerful artificial intelligence (AI) tool for handling complex tasks. However, \dl typically requires extensive training data to provide robust results \cite{lit:robustness}. High acquisition costs can render the collection of sufficient data unfeasible. This is especially problematic in safety-critical domains like autonomous driving, where we encounter a wide range of edge cases associated with high risks \cite{lit:kiwissen}. Informed learning (IL) aims to improve the data efficiency and robustness of \dl models by integrating prior knowledge~\cite{lit:informed}. Most \il approaches consider prior scientific knowledge by constraining or verifying the problem space or learning process directly. However, hard constraints are not suitable for qualitative prior expert and world knowledge where ubiquitous exceptions exist. In autonomous driving, for example, we expect traffic participants to comply with speed regulations but must not rule out violations. Still, prior knowledge about norms and regulations, like in this example, are highly informative for most cases and readily available at low costs.

A recent idea is the integration of such prior expert and world knowledge into probabilistic \dl models~\cite{lit:bayesian_transfer,lit:schlauch}. These models maintain a distribution over possible model parameters instead of single maximum likelihood estimates. The prior knowledge can be represented as a prior parameter distribution, learned from arbitrarily defined knowledge tasks, to regularize training on real-world observations. The probabilistic informed learning (PIL) approach of Schlauch~\shortcite{lit:schlauch} applies this idea to the trajectory prediction in autonomous driving using regularization-based continual learning methods, achieving a substantially improved data efficiency. However, typical sampling-based probabilistic \dl model approximations, such as the variational inference (VI) used by Schlauch~\shortcite{lit:schlauch}, are computationally expensive, since they require multiple inference passes and substantially more training epochs. A promising alternative are compute-efficient last layer approximations~\cite{lit:last_layer}. The spectral normalized Gaussian process (SNGP)~\cite{lit:sngp} is a particularly efficient approximation, that applies a Gaussian process (GP) as last layer to a deterministic deep neural network (DNN). The DNN acts as scalable feature extractor, while the last layer GP allows the deterministic estimation of the uncertainty in a single inference pass. The last layer GP kernel itself is approximated via Fourier features, which is asymptotically exact and can be easily scaled. 

We propose a novel regularization-based continual learning method to enable the use of SNGPs in a \pil approach. Our proposal is conceptually simple, builds upon well-established methods \cite{lit:online_ewc,lit:gvcl}, imposes little computational overhead and requires no additional architecture changes, making implementation straightforward. We apply our method in a \pil approach for the trajectory prediction in autonomous driving, which is an especially challenging application since well-calibrated, multi-modal predictions are required to enable safe planning. 

\begin{figure*}[t]
\includegraphics[width=\linewidth]{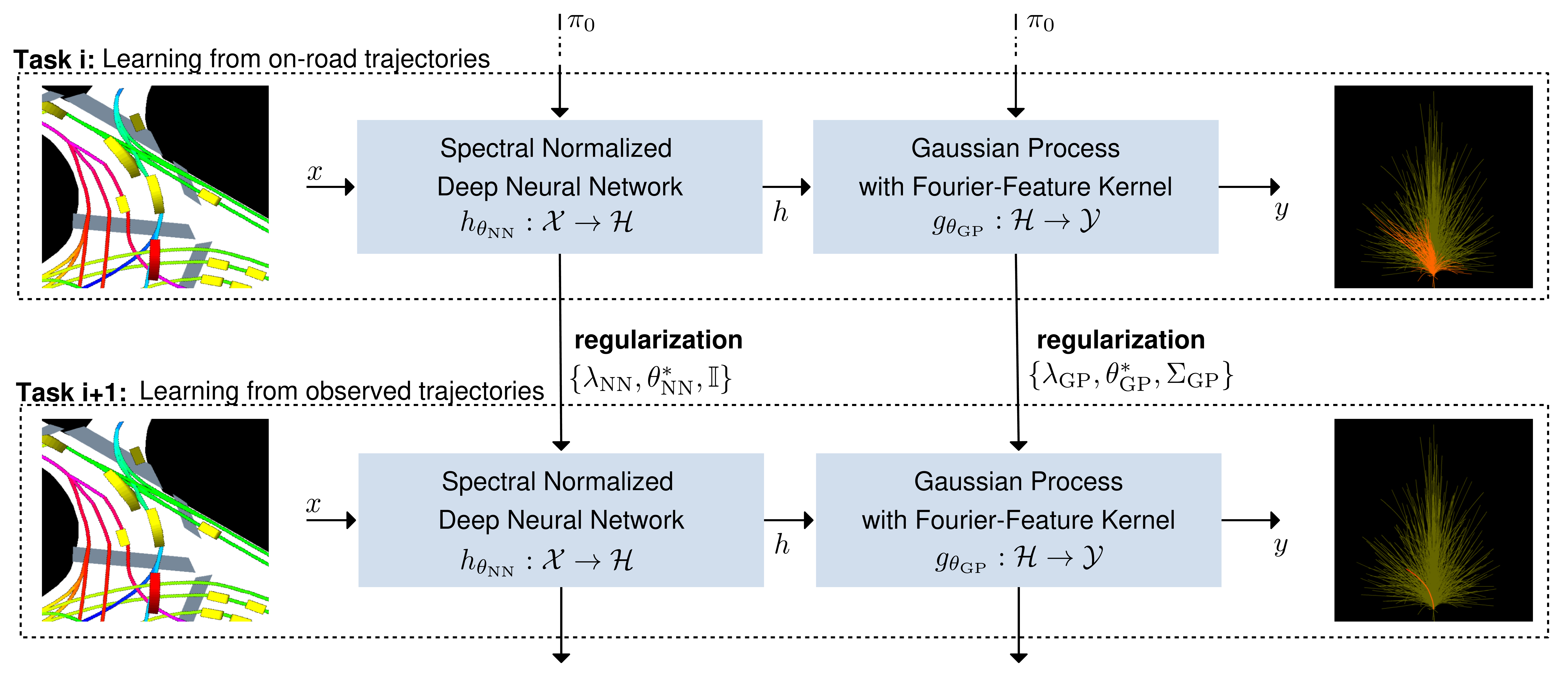}
\caption{The informed CoverNet-SNGP model consists of a spectral normalized feature extractor and a last layer Gaussian Process with a Fourier feature approximated radial basis function kernel. Given a Birds-Eye-View RGB rendering and the target's current state, the model classifies a set of candidate trajectories according to their drivability in task $i$ and their likely realization in task $i+1$. Our method regularizes the training on task $i+1$, given the MAP estimates and Laplace approximated covariance from task $i$ as informative priors, thereby integrating the drivability knowledge following the PIL approach.}
\label{fig:flowchart}
\end{figure*}

Following Schlauch~\shortcite{lit:schlauch}, we employ CoverNet as base model and integrate the prior \textit{drivability} knowledge that trajectories are likely to stay on-road. We benchmark our proposed \textit{informed CoverNet-SNGP} on two public datasets, NuScenes and Argoverse2, against the non-informed Base-CoverNet, CoverNet-SNGP and informed Transfer-CoverNet, GVCL-Det-CoverNet as baselines. To this end, we evaluate data-efficiency under diminishing training data availability and robustness to location-transfers, both being key aspects for safe autonomous driving~\cite{lit:shifts,lit:kiwissen}. We observe benefits in favor of our informed CoverNet-SNGP across various performance metrics, especially in low data regimes, which demonstrates our method's viability to increase data-efficiency and robustness in a \pil approach. Our code is available on GitHub\footnote{https://github.com/continental/kiwissen-bayesian-trajectory-prediction}.

\section{Related Work}
\label{sec:related_works}

Van Rueden~\shortcite{lit:informed} provides an overview of \il as an emerging field of research, which is also known as knowledge-guided or -augmented learning \cite{lit:kiwissen}. In trajectory prediction, like in other domains, most work concentrates on integrating prior scientific knowledge. Dynamical models are used, for instance, to encode physical limitations of motion in the architecture \cite{lit:kinematic}, in the output representation \cite{lit:covernet} or in a post-hoc verification \cite{lit:mpc}. Approaches similar to the \pil approach~\cite{lit:schlauch}, that focus on integrating expert and world prior knowledge, typically leverage transfer- or multi-task learning settings \cite{lit:transfer_baseline}. However, transfer learning does not prevent catastrophic forgetting, while multi-task learning requires a single dataset with simultaneously available labels. \pil can be applied without these limitations. 

SNGPs and related models, known as deterministic uncertainty models (DUMs), have been analyzed by Postels~\shortcite{lit:practicality_dums} and Charpentier~\shortcite{lit:training_dums}. Most closely related to SNGPs is the deterministic uncertainty estimator (DUE) proposed by van~Amersfoort~\shortcite{lit:due}, which approximates the last layer kernel with sparse variational inducing points instead of Fourier features. DUE preserves the non-parametric nature of the kernel, but is sensitive to its initialization and generally not asymptotically exact. 

Parisi~\shortcite{lit:cl_review} and De Lange~\shortcite{lit:cl_survey} give a detailed survey of continual learning methods and their classification. Our proposed continual learning method for SNGPs is purely regularization-based, in contrast to the functional regularization introduced by Titsias~\shortcite{lit:frcl}, which could be directly applied to the DUE model, and the work of Derakhshani~\shortcite{lit:kcl}, which also considers a kernel approximation based on Fourier features. Both these methods require rehearsal, the latter also a parameter expansion. Rehearsal is likely to be sensitive to the data imbalances \cite{lit:replay} in our application, while parameter expansions require architecture changes which introduce additional complexity. Our proposed method is conceptually simple and builds upon the well-established online elastic weight consolidation (online EWC) introduced by Schwartz~\shortcite{lit:online_ewc}. Online EWC can also be understood as special case of generalized variational continual learning (GVCL) described by Loo~\shortcite{lit:gvcl}.

\section{Informed SNGPs}
\label{sec:method}

\subsection{Probabilistic Informed Learning}
\label{sec:pil}

The \pil approach of Schlauch~\shortcite{lit:schlauch} integrates prior expert and world knowledge in a supervised learning setup. The basic idea is to define a sequence of {knowledge} tasks $i=1,\ldots,M-1$ on datasets $D_i=\{(x^{(i)}_j, y^{(t)}_j)\}^{n_i}_{j=1}$ with $n_i$ samples each. These datasets can be synthetically generated, for example, by leveraging semantic annotations to map the prior knowledge to the prediction target. Semantic annotations are readily available in domains like autonomous driving, but are often underutilized in state-of-the-art models that learn from observations in the conventional task $i=M$ alone \cite{lit:walk_alone}.

Given a probabilistic \dl model parameterized by $\theta$ and an initial uninformative prior $\pi_0(\theta)$, the goal is to recursively learn from the sequence of tasks by applying Bayes' rule
\begin{equation}\label{eq:Bayes}
\begin{split}
    p(\theta| D_{1:i}) \propto \pi_0(\theta) \prod^i_{j=1}p_{\theta}(y_j|x_j),
\end{split}
\end{equation}
where $p_{\theta}(y_j|x_j)$ are the likelihood functions at task $j$, which are assumed to be conditionally independent given $\theta$. This computationally intractable recursion is approximated by repurposing regularization-based continual learning methods. 

The \pil approach can generally be applied, as long as first, the prior knowledge is strongly related to the observational task, second, the prior knowledge can be mapped to the prediction target and third, the posterior parameter distribution can be estimated. The informative priors make information explicit and shape the loss surface in the downstream task, improving the training outcome; even without using probabilistic inference in the end~\cite{lit:bayesian_transfer}.

\subsection{SNGP Composition}
\label{sec:sngp}

SNGPs \cite{lit:sngp} employ a composition $f_\theta=g_{\theta_{\text{GP}}} \circ h_{\theta_{\text{NN}}}: \mathcal{X} \to \mathcal{Y}$, $\theta=\{\theta_\text{NN}, \theta_{\text{GP}}\}$. Its first component is a deterministic, spectral normalized feature extractor $h_{\theta_{\text{NN}}}: \mathcal{X} \to \mathcal{H}$ with trainable parameters $\theta_{\text{NN}}$ mapping the high dimensional input space $\mathcal{X}$ into a low dimensional hidden space $\mathcal{H}$. 
The second component is a GP output layer $g_{\theta_{\text{GP}}}: \mathcal{H} \to \mathcal{Y}$ with a radial basis function (RBF) kernel mapping into the output space $\mathcal{Y}$. The RBF kernel can be approximated by (random) Fourier features using Bochner's Theorem \cite{lit:bochner_theorem}. This effectively reduces the GP to a Bayesian linear model, that can be written as a neural network layer with fixed hidden weights and trainable output weight parameters $\theta_{\text{GP}}$ and enables end-to-end training with the feature extractor. The distance-sensitive of the composition prevents a ``feature-collapse'' \cite{lit:due}, improving the calibration against adversarial and outlier samples. In total, SNGP introduces five additional hyperparameters, namely an upper bound $s$ and number of power iterations $N_p$ for the spectral normalization for the feature extractor and the number of Fourier features $N_f$, the kernel's length scale $l_s$ and Gaussian prior choice for the last layer.

\subsection{Regularizing SNGPs}
\label{sec:sngp_reg}

There are two problems prohibiting the direct application of the  \pil approach to composite last layer kernel approximations like the SNGP. First, there is no existing continual learning method for kernels that does not require rehearsal memories or parameter expansions (see Sec.~\ref{sec:related_works}). Second, estimating the posterior parameter distribution of the feature extractor (e.g.~via a Laplace approximation or variational inference) contradicts the motivation for the last layer kernel approximation regarding compute-efficiency.

We tackle the first problem by leveraging the Fourier feature approximation of the  RBF kernel of the GP. The posterior distributions of the parameters of the last layer at task $i$ can be made tractable through Laplace approximation \cite{lit:sngp}, that is, we assume
\begin{align*}
    p(\theta_\text{GP}| D_{1:i})  \approx \mathcal{N}(\theta_\text{GP}; \theta_{\text{GP}, i}^{*}, \Sigma^{-1}_{\text{GP},i}),
\end{align*}
given a maximum a posteriori (MAP) estimate $\theta^*_{\text{GP}, i}$ at task $i$. Similar to online EWC \cite{lit:online_ewc},  $\theta^*_{\text{GP}, i}$ can be obtained by minimizing
\begin{equation}
\label{eq:gp_map}
            \displaystyle
    -\log{p_{\theta_{\text{GP}}}(y_{i} | x_{i})} - \frac{\lambda_\text{GP}}{2}(\theta_{\text{GP}} - \theta^*_{\text{GP}, i-1})^\top \Sigma_{\text{GP}, i-1}^{-1}(\theta_{\text{GP}} - \theta^*_{\text{GP}, i-1})
\end{equation}
with respect to $\theta_{\text{GP}}$, where the precision $\Sigma_{\text{GP}, i}^{-1}$ is approximated by the sum of the Hessian at the MAP estimate and a scaled precision at task $i-1$, that is,
\begin{equation*}
   \Sigma_{\text{GP}, i}^{-1} \approx H_{\text{GP}, i}(\theta_{\text{GP}, i}^{*}) + \gamma_\text{GP} \Sigma_{\text{GP}, i-1}^{-1}.
\end{equation*}
Above, $\lambda_\text{GP}>0$ is a temperature parameter, that scales the importance of the previous task \cite{lit:ewc}, and $0<\gamma_\text{GP}\leq 1 $ is a decay parameter, that allows for more plasticity over very long task sequences \cite{lit:online_ewc}. In contrast to online EWC, we can cheaply compute the Hessian using moving averages~\cite{lit:sngp} instead of using a Fisher matrix approximation. In the first task $i=1$, we use an uninformative zero-mean, unit-variance prior $\pi_0$, which amounts to a simple $\mathcal{L}$2-regularization. 

To tackle the second problem and regularize the feature extractor, we approximate the precision $\Sigma_{\text{NN},i-1}^{-1}$ with the identity matrix $\mathbb{I}$. This implies a simple $\mathcal{L}$2-regularization for the MAP estimates $\theta^*_{\text{NN}, i}$ obtained by minimizing
\begin{align*}
    &-\log{p_{\theta_\text{NN}}(y_{i} | x_{i})} - \frac{\lambda_\text{NN}}{2}(\theta_\text{NN} - \theta^*_{\text{NN}, i-1})^2
\end{align*}
with respect to $\theta_\text{NN}$, where $\lambda_\text{NN}$ is the extractor specific temperature parameter. This idea is conceptually simple, but should be sufficient, since the learned representation in knowledge tasks should be suitable downstream due to the close relation between tasks. 

In result, the complete model $f_{\theta}: \mathcal{X} \to \mathcal{Y}$, parameterized by $\theta=\{\theta_\text{NN}, \theta_\text{GP}\}$, can be effectively regularized and used in the \pil approach, as visualized in Fig.~\ref{fig:flowchart}. Our method introduces three hyperparameters $\{\lambda_\text{GP}, \gamma_\text{GP} , \lambda_\text{NN}\}$. It only requires the   parameters of the previous task in memory and has little computational overhead like online EWC \cite{lit:online_ewc}.

\section{Application to Trajectory Prediction}
\label{sec:application}

\subsection{Problem Definition}

We limit ourselves to the \textit{single-agent} trajectory prediction problem \cite{lit:tp_survey}. An autonomous driving system is assumed to observe the states in the state space $\mathcal{Y}$ of all agents $\mathcal{A}$ present in a scene on the road. Let $y^{(t)} \in \mathcal{Y}$ denote the state of target agent $a \in \mathcal{A}$ at time $t$ and let $y^{(t-T_o\,:\,t)}=\big( y^{(t-T_o)}, y^{(t-T_o+\delta t)}, \ldots, y^{(t)}\big)$ be its observed trajectory over an observation history $T_o$ with sampling period $\delta t$. Additionally, we assume access to agent-centered maps $\mathcal{M}$, which include semantic annotations such as the drivable area. Map and states make up the scene context of agent $a$, denoted as ${x=(\{y_j^{(t-T_o\,:\,t)}\}^{|\mathcal{A}|}_{j=1},\mathcal{M})}$. Given $x$, the goal is to predict the distribution of $a$'s future trajectories $p(y^{(t+\delta t\,:\,t+T_h)}|x)$ over the prediction horizon $T_h$, where $y^{(t-\delta t\,:\,t+T_h)}=\big( y^{(t+\delta t)}, y^{(t+2\delta t)}, \ldots, y^{(t+T_h)}\big)$.

\subsection{CoverNet-SNGP}

CoverNet \cite{lit:covernet} approaches the single-agent trajectory problem by considering a birds-eye-view RGB rendering of the scene context $x$ and the current state $y^{(t)}$ of the target agent $a$ as inputs. The RGB rendering is processed by a computer-vision backbone, before concatenated with the target's current state and processed by another dense layer. The output is represented as a set $\mathcal{K}$ of $K$ candidate trajectories $y_{k}^{(t+\delta t\,:\,t+T_h)}$. Doing so reduces the prediction problem to a classification problem, where each trajectory  in the set $\mathcal{K}$ is treated as a sample of the predictive distribution $p(y^{(t+\delta t\,:\,t+T_h)}|x)$ and only the conditional probability of each sample is required. In principle, any space-filling heuristic may be used to define $\mathcal{K}$, for example, by using a dynamical model that integrates physical limitations \cite{lit:covernet}, which could be applied in combination with the \pil approach. Here, we follow Phan-Minh's~\shortcite{lit:covernet} definition of a fixed set $\mathcal{K}$ by solving a set-covering problem over a subsample of observed trajectories in the training split, using a greedy-algorithm\footnote{Further details in our supplemental. Also see Chapter 35.3 of Cormen~\shortcite{lit:algo_data} regarding set-covering problems in general.} given a coverage-bound $\epsilon$, which determines the number of total candidates $K$.

The modification of CoverNet with SNGP is straightforward if a convolutional neural network (CNN) is used as backbone. In that case, a spectral normalization can be directly applied to the architecture while the last layer is replaced with a Gaussian process, approximated by Fourier features as described in Sec.~\ref{sec:sngp}. 

\subsection{Integrating Prior Drivability Knowledge}
\label{sec:tasks}

The \pil approach is applied sequentially on two consecutive tasks as follows. In task $i$, we integrate the prior drivability knowledge, that trajectories are likely to stay on-road. To this end, we derive new training labels (see Sec.~\ref{sec:pil}), where all candidate trajectories in $\mathcal{K}$ with way-points inside the drivable area for a given training scene $x$ are labeled as positive~\cite{lit:transfer_baseline}. We then train in a multi-label classification with a binary cross-entropy loss on these labels. In  task $i+1$, the closest candidate trajectory in $\mathcal{K}$ to the observed ground truth is labeled as positive. We train in a multi-class classification with a sparse categorical cross-entropy loss (using softmax normalized logit transformations) on these labels \cite{lit:covernet}. In effect, the consecutive tasks are only differing in the labels and loss functions used. Applying our method described in Sec.~\ref{sec:sngp_reg}, we first train our CoverNet-SNGP model on task $i$ and then regularize its training on task $i+1$, as exemplified in Fig.~\ref{fig:flowchart}. We denote the resulting informed CoverNet-SNGP as {CoverNet-SNGP}$_\textbf{I}$, opposed to the non-informed version {CoverNet-SNGP}$_\textbf{U}$ trained on task $i+1$ only without integration of prior knowledge from task $i$.

\section{Experimental Design}

\subsection{Datasets}

We use the public NuScenes \cite{lit:nuscenes} and Argoverse2 \cite{lit:argo} datasets. We replicate the NuScenes data split by Phan-Minh~\shortcite{lit:covernet} on Argoverse2, only considering vehicle targets (exlcuding pedestrians and cyclists not driving on-road), as summarized in Tab.~\ref{tab:splits}. For the RGB rendering, we consider each scene with a one-second history ($T_o=1\text{s}$). For the candidate trajectories in $\mathcal{K}$, we consider a six-second prediction horizon ($T_h=6\text{s}$), sampled at $2\text{Hz}$ in NuScenes and $10\text{Hz}$ in Argoverse2. Both datasets include drivable areas in the semantic map data, allowing us to define the first task as described in Sec.~\ref{sec:tasks}.

\begin{table}[h]
\tiny
\caption{Numbers and percentages of samples across location subsets of both NuScenes and Argoverse2.} \label{tab:splits}
\begin{center}
\begin{tabular}{l*{3}{r}}
\toprule
\textbf{data subset} & train split \# (\%)  & train-val split \# (\%)  & val split \# (\%) \\
\midrule
NuScenes Total & 32186 (100.0) &  8560 (100.0) & 9041 (100.0) \\
\hspace*{3mm} Boston       & 19629 (60.99)   &   5855 (68.40)  & 5138 (56.84) \\ 
\hspace*{3mm} Singapore       & 12557 (49.01)  & 2705 (31.60)     & 3903 (43.16)\\
\midrule
Argoverse2 Total & 161379 (100.0) &  22992 (100.0) &  23113 (100.0) \\
\hspace*{3mm} Miami       &  42214 (26.16)  &  5983  (26.02)   &  5984 (25.89) \\
\hspace*{3mm} Austin      &  34681 (21.49)  &  4968 (21.57)   & 4985 (26.16) \\
\hspace*{3mm} Pittsburgh      &  33391 (20.69) &  4823  (20.98)   & 4803 (20.78)\\
\hspace*{3mm} Dearborn      &  20579  (12.75) &  2933 (12.79)   & 3001 (12.98)\\
\hspace*{3mm} Washington-DC      &  20546 (12.73) &  2883 (12.54) & 2976 (12.88) \\
\hspace*{3mm} Palo-Alto       &  9968  (6.18) &  1402 (6.10)   & 1364 (5.90)\\
\bottomrule
\end{tabular}
\end{center}
\end{table}

\subsection{Baselines}

We consider the unmodified CoverNet as baseline, once as non-informed {Base-CoverNet}~\cite{lit:covernet} and once as {Transfer-CoverNet}. The Transfer-CoverNet baseline, pretrained on task $i$ and then trained on the current task $i+1$, has previously been proposed by Boulton~\shortcite{lit:transfer_baseline}. We can also understand it as an ablation-type baseline to the \pil approach without regularization. In addition, we compare to {GVCL-Det-CoverNet} proposed by Schlauch~\shortcite{lit:schlauch}, since it only needs a single-inference pass too. However, GVCl-Det-CoverNet also requires computationally extremely expensive training of a GVCL-CoverNet model. For example, in our setting, training until convergence on a single Nvidia RTX A5000 GPUs with $10\%$ of NuScenes data needs around 120 hours for GVCL-CoverNet, in contrast to 8 hours for {CoverNet-SNGP}$_\textbf{I}$ and 6 hours for Base-CoverNet.

\subsection{Metrics}

We measure the average displacement error \ade[n]{1} and final displacement error \fde[n]{1}, evaluating the quality of the most likely trajectory, and the \ade[n]{5}, which considers the five most likely trajectories \cite{lit:tp_survey}. The \ade[n]{5} depends on the probability-based ordering and, thus, indirectly on the calibration. We also consider the drivable area compliance (DAC) to evaluate the extent to which predictions align with our prior drivability knowledge.

Since observed ground truth trajectories may not be part of the trajectory set $y_{\text{true}}^{(t+\delta t\, :\, t+T_h)} \notin \mathcal{K}$, the CoverNet model exhibits an irreducible approximation error. To more clearly assess the impact of our method, we also consider the classification-based negative log likelihood (NLL) and the rank of the positively labeled trajectory (RNK), both directly depending on the calibration, and the Top1-accuracy (ACC).

\subsection{Implementation Details}\label{sec:imp}

We use the output representation described in Sec.~\ref{sec:application} with a coverage bound $\epsilon = 4 \text{m}$, for NuScenes with $K_{\text{Nusc}}=415$ and for Argoverse2 with $K_{\text{Argo}}=518$ candidates. We employ a ResNet-50 as backbone and SGD as optimizer. For the CoverNet-SNGPs, we fix power iterations $N_p$ to one and the number of Fourier features $N_f$ to 1024, following Liu~\shortcite{lit:sngp}. The spectral normalization's upper bound $s$ and the kernel length scale $l_s$ are treated as additional hyperparameters. We tune the hyperparameters of each model on the respective tasks with 100\% of the data using the validation NLL\footnote{Configurations are available in our supplemental and on Github.}. The exception is {CoverNet-SNGP}$_\textbf{I}$, which uses the same settings as {CoverNet-SNGP}$_\textbf{U}$ on task $i+1$. We also fix both temperature parameters $\lambda_\text{NN}$ and $\lambda_\text{GP}$ ad-hoc to the inverse of the effective dataset size to keep tuning costs low. The decay parameter $\gamma_\text{GP}$ is mostly relevant for very long task sequences (see Sec.~\ref{sec:method}), such that we set $\gamma_\text{GP}=1$.

\section{Results}

We study the performance of our CoverNet-SNGP$_\text{I}$ against the baselines under two sets of experiments. First, we investigate the performance under increasingly smaller subsets of the observational training data, allowing us to shed light on data-efficiency. These subsets are randomly subsampled once and then kept fixed across models and repetitions. In this set, we also consider GVCL-Det-CoverNet with results on NuScenes for $100\%$, reported from Schlauch~\shortcite{lit:schlauch}, $10\%$ and $3\%$, replicated with only three independent repetitions, due to the long training times. Second, we test the performance by training and testing on location-specific subsets, gaining insights into the robustness to location-transfers, which is often implicitly assumed in the state of the art \cite{lit:shifts}. The reported results are the average performance and standard deviation of five independent runs for each experiment.  

\subsection{Effect of Available Training Data}
\label{sec:res_data}

\begin{figure*}
\tiny
\centering
\begin{minipage}[c]{\linewidth}
\captionof{table}{Average performance and standard deviation of 5 independent repetitions over decreasing subsamples of NuScenes (\textbf{bold} as best).} 
\label{tab:data_nuscenes}
\vspace{-10pt}
\begin{center}
\begin{tabular}{rl*{7}{R}}
\toprule
\textbf{Data} (in \%)& \textbf{Model} & \ade[b]{1} & \ade[b]{5}& \fde[b]{1}  & \textbf{NLL} & \textbf{RNK} & \textbf{ACC} (in \%) &  \textbf{DAC} (in \%)\\
\midrule
100 & Base   & 4.92 \tpm 0.15& 2.34 \tpm 0.05& 10.94 \tpm 0.27& 3.47 \tpm 0.06& 15.55 \tpm 0.73 & 13.94 \tpm 1.10 &  89.26 \tpm 1.13\\
& Transfer & 4.60 \tpm 0.04& \textbf{2.18 \tpm 0.02} & 9.94 \tpm 0.08 & 3.21 \tpm 0.01& \textbf{11.79 \tpm 0.18} & 15.19 \tpm 0.43 & \textbf{95.73 \tpm 0.29} \\
& GVCL-Det* & 4.55 \tpm 0.11 & 2.26 \tpm 0.05 & \textbf{9.93 \tpm 0.39} & 3.60 \tpm 0.08& 11.85 \tpm0.48 &14.88 \tpm0.94 & 90.94 \tpm2.25 \\
& SNGP$_\text{U}$   & 4.53 \tpm 0.09 & 2.25 \tpm 0.04& 10.31  \tpm 0.27 & 3.23 \tpm 0.01& 13.25 \tpm 0.19 & 17.04 \tpm 0.68 & 91.19 \tpm 0.61\\
& SNGP$_\text{I}$   & \textbf{4.45 \tpm 0.04}& 2.21 \tpm 0.01& 10.09  \tpm 0.12& \textbf{3.19 \tpm 0.01}& 12.44 \tpm 0.14 & \textbf{17.36 \tpm 0.59} & 91.65 \tpm 0.59\\
  \rule{0pt}{2.2ex}
50 & Base   &5.15 \tpm 0.23 & 2.37 \tpm 0.11& 11.46 \tpm 0.60 &3.52 \tpm 0.06 & 17.21 \tpm 1.33 & 13.55 \tpm 0.62 &  86.68 \tpm 4.72\\
& Transfer &4.86 \tpm 0.04 & 2.26 \tpm 0.01& 10.38 \tpm 0.06& 3.35 \tpm 0.01& 13.46 \tpm 0.21 & 14.37 \tpm 0.09&  \textbf{95.66 \tpm 0.28}\\
&SNGP$_\text{U}$   & 4.57 \tpm 0.05& 2.26 \tpm 0.04 & 10.40  \tpm 0.15& 3.30  \tpm 0.02& 14.62 \tpm 0.17& \textbf{16.83 \tpm 0.59} & 90.09 \tpm 0.56\\
& SNGP$_\text{I}$   & \textbf{4.48 \tpm 0.07} & \textbf{2.22 \tpm 0.04}& \textbf{10.13  \tpm 0.13} & \textbf{3.25  \tpm 0.02}&  \textbf{13.39\tpm 0.31} & 16.72 \tpm 0.76 & 91.10 \tpm 0.72\\
  \rule{0pt}{2.2ex}
30 & Base   & 5.40 \tpm 0.03& 2.44 \tpm 0.07&12.01  \tpm 0.20 & 3.68  \tpm 0.04& 19.80 \tpm 1.03 & 12.70 \tpm 0.81 &  86.58 \tpm 2.54\\
& Transfer & 5.08 \tpm 0.03& 2.34 \tpm 0.02& 10.80  \tpm 0.07& 3.47 \tpm 0.01& 15.00 \tpm 0.05 & 13.38 \tpm 0.31 &  \textbf{96.07 \tpm 0.32}\\
& SNGP$_\text{U}$   & 4.68 \tpm 0.09& \textbf{2.29 \tpm 0.04} & 10.61  \tpm 0.26& 3.37 \tpm 0.01& 16.02 \tpm 0.22 & 16.69 \tpm 0.62& 89.22 \tpm 0.30\\
& SNGP$_\text{I}$   & \textbf{4.58 \tpm 0.03} & 2.30 \tpm 0.02 & \textbf{10.35  \tpm 0.08} & \textbf{3.31  \tpm 0.02} & \textbf{14.67 \tpm 0.20} & \textbf{17.10 \tpm 0.34} & 90.41 \tpm 0.49\\
  \rule{0pt}{2.2ex}
10 & Base   & 5.89 \tpm 0.28& 2.72 \tpm 0.11&12.88  \tpm 0.63 & 3.99  \tpm 0.06& 32.74 \tpm 1.48 & 12.38 \tpm 0.96 &  86.38 \tpm 2.64\\
& Transfer & 6.09 \tpm 0.03& 2.65 \tpm 0.02& 12.60  \tpm 0.06& 3.89 \tpm 0.01& 24.82 \tpm 0.13 & 10.35 \tpm 0.15&  \textbf{95.54 \tpm 0.23}\\
& GVCL-Det* & 5.27 \tpm 0.27 & 2.53 \tpm  0.09 & 12.03 \tpm  0.58 & 4.05 \tpm 0.07 & 24.78 \tpm 0.45& 12.95 \tpm 0.80 & 91.52 \tpm  1.54 \\
& SNGP$_\text{U}$   & 5.00 \tpm 0.04& 2.52 \tpm 0.03& 11.36 \tpm 0.22& 3.60 \tpm 0.02& 25.19 \tpm 0.30 & \textbf{15.73 \tpm 0.22} & 88.62 \tpm 0.56\\
& SNGP$_\text{I}$   & \textbf{4.96 \tpm 0.05} & \textbf{2.47 \tpm 0.04} & \textbf{11.25  \tpm 0.16} & \textbf{3.52 \tpm 0.03} & \textbf{20.94 \tpm 0.59} & 15.39 \tpm 0.29 & 89.53 \tpm 1.11\\
  \rule{0pt}{2.2ex}
5 & Base   & 5.90 \tpm 0.17& 2.82 \tpm 0.06& 12.81 \tpm 0.38& 4.26 \tpm 0.03& 42.55 \tpm 1.92  & 10.17 \tpm 1.26 &  86.89 \tpm 1.96\\
& Transfer & 6.62 \tpm 0.04& 2.89 \tpm 0.01 & 13.41 \tpm 0.09& 4.30 \tpm 0.01& 29.74 \tpm 0.44 & 8.70 \tpm 0.14 &  \textbf{97.46 \tpm 0.07}\\
& SNGP$_\text{U}$   & 5.07 \tpm 0.05 & 2.58 \tpm 0.02& 11.63 \tpm 0.13 & 3.90 \tpm 0.03& 31.77 \tpm 0.85 & 14.29 \tpm 0.30 & 86.31 \tpm 0.82\\\
& SNGP$_\text{I}$   & \textbf{5.01 \tpm 0.04} & \textbf{2.53 \tpm 0.04} & \textbf{11.43 \tpm 0.10} & \textbf{3.72 \tpm 0.03} & \textbf{25.99 \tpm 0.65} & \textbf{14.32 \tpm 0.59} & 86.85 \tpm 0.94\\
  \rule{0pt}{2.2ex}
3 & Base   & 6.23 \tpm 0.16& 3.11 \tpm 0.11& 13.32  \tpm 0.28& 4.53  \tpm 0.03& 59.34 \tpm 3.76 & 10.42 \tpm 0.71 & 84.83 \tpm 2.00 \\
& Transfer & 7.52 \tpm 0.09& 3.35 \tpm 0.07& 14.71 \tpm 0.14& 4.61 \tpm 0.01& 36.62 \tpm 0.60 & 7.33 \tpm 0.10 &  \textbf{97.80 \tpm 0.08}\\
& GVCL-Det* &  6.12 \tpm 0.11& 2.86 \tpm 0.09& 13.25 \tpm 0.31 & 4.26 \tpm 0.05& 31.96 \tpm 3.01& 10.87 \tpm 0.49& 93.05 \tpm 1.21\\
& SNGP$_\text{U}$   & 5.56 \tpm 0.09 &  2.85 \tpm 0.07&12.64  \tpm 0.15 &4.61  \tpm 0.02 & 46.82 \tpm 1.37 & \textbf{13.37 \tpm 0.09} & 86.00 \tpm 0.95\\
& SNGP$_\text{I}$   & \textbf{5.44 \tpm 0.13} & \textbf{2.74 \tpm 0.06} & \textbf{12.38 \tpm 0.29} & \textbf{3.90 \tpm 0.01} & \textbf{27.88 \tpm 1.54} & 12.62 \tpm 0.64 & 86.18 \tpm 0.77\\
  \rule{0pt}{2.2ex}
1 & Base   & 8.39 \tpm 1.16& 3.44 \tpm 0.30&16.25 \tpm 2.27 & 5.23  \tpm 0.10& 83.20 \tpm 3.84 & 5.39 \tpm 1.92 &  81.48 \tpm 4.34\\
& Transfer & 8.44 \tpm 0.07& 4.18 \tpm 0.08 &15.71 \tpm 0.27 & 5.52  \tpm 0.01& 52.92 \tpm 0.11 & 4.53 \tpm 0.15 &  \textbf{98.29 \tpm 0.27}\\
& SNGP$_\text{U}$   & 6.33 \tpm 0.64& 2.88 \tpm 0.07 &12.76 \tpm 1.07 & 5.48  \tpm 0.01& 77.64 \tpm 3.38 & 8.48 \tpm 0.92 & 70.42 \tpm 3.97\\
& SNGP$_\text{I}$   & \textbf{5.39 \tpm 0.28} & \textbf{2.68 \tpm 0.07} &\textbf{12.27 \tpm 0.53} & \textbf{4.40  \tpm 0.01} & \textbf{50.19 \tpm 1.15} & \textbf{9.94 \tpm 0.56} & 79.34 \tpm 2.54\\
\bottomrule
\end{tabular}
\end{center}
\vspace{0.1mm}
\end{minipage}
\begin{minipage}[c]{\linewidth}
\captionof{table}{Average performance and standard deviation of 5 independent repetitions over decreasing subsamples of Argoverse2 (\textbf{bold} as best).} \label{tab:data_argo}
\vspace{-10pt}
\begin{center}
\begin{tabular}{rl*{7}{R}}
\toprule
\textbf{Data} (in \%) & \textbf{Model} & \ade[b]{1} & \ade[b]{5}& \fde[b]{1}  & \textbf{NLL} & \textbf{RNK} & \textbf{ACC} (in \%) & 
 \textbf{DAC} (in \%) \\
\midrule
100 & Base   & 3.57 \tpm 0.07 & 1.84 \tpm 0.04& 8.96 \tpm 0.15 & 2.73\tpm 0.01& 7.87 \tpm 0.58 & 24.46 \tpm 0.59 &  94.77 \tpm 0.32\\
& Transfer & 3.60 \tpm 0.04& \textbf{1.76 \tpm 0.02} & 8.78 \tpm 0.08 & \textbf{2.68 \tpm 0.01} & \textbf{7.31 \tpm 0.18} & 24.61 \tpm 0.33 & \textbf{96.91 \tpm 0.09} \\
& SNGP$_\text{U}$   & 3.60 \tpm 0.04 & 1.86 \tpm 0.03& 9.00  \tpm 0.15 & 2.74 \tpm 0.03& 8.19 \tpm 0.28 & 25.24 \tpm 0.29 & 95.00 \tpm 0.47\\
& SNGP$_\text{I}$   & \textbf{3.51 \tpm 0.06} & 1.82 \tpm 0.03& \textbf{8.73  \tpm 0.07} & 2.69 \tpm 0.01& 7.69 \tpm 0.17 & \textbf{25.56 \tpm 0.42} & 95.01 \tpm 0.12 \\
  \rule{0pt}{2.2ex}
50 & Base   &3.93 \tpm 0.10 & 1.97 \tpm 0.07& 9.78 \tpm 0.31 &2.98 \tpm 0.07 &9.89 \tpm 0.53 & 21.02 \tpm 1.17 &  93.99 \tpm 1.58\\
& Transfer &3.80  \tpm 0.01 & \textbf{1.83 \tpm 0.02} & \textbf{9.36 \tpm 0.05} & \textbf{2.80 \tpm 0.01} & \textbf{8.16 \tpm 0.02} & 23.41 \tpm 0.29 &  \textbf{97.04 \tpm 0.22}\\
& SNGP$_\text{U}$   & 3.84 \tpm 0.04 & 2.01 \tpm 0.02 & 9.67  \tpm 0.15& 2.89  \tpm 0.02& 9.95 \tpm 0.28 & 23.53 \tpm 0.40 & 94.93 \tpm 0.19\\
& SNGP$_\text{I}$   & \textbf{3.76 \tpm 0.02} & 1.95 \tpm 0.02& 9.38  \tpm 0.06 & 2.84  \tpm 0.02&  9.27 \tpm 0.15 & \textbf{23.81 \tpm 0.76} & 94.92 \tpm 0.64\\
  \rule{0pt}{2.2ex}
30 & Base   &4.22 \tpm 0.10 & 2.04 \tpm 0.01& 10.41 \tpm 0.28 &3.07 \tpm 0.06 &11.18 \tpm 1.33 & 19.76 \tpm 0.49 &  94.37 \tpm 0.71\\
& Transfer & 3.99 \tpm 0.02& \textbf{1.89 \tpm 0.02} & \textbf{9.76 \tpm 0.05} & \textbf{2.91 \tpm 0.01} & \textbf{9.07 \tpm 0.04} & 22.02 \tpm 0.23 & \textbf{97.15 \tpm 0.23} \\
& SNGP$_\text{U}$   & 3.98 \tpm 0.03& 2.09 \tpm 0.04& 9.95  \tpm 0.09& 2.99 \tpm 0.01& 11.40 \tpm 0.30 & \textbf{22.79 \tpm 0.22} & 94.73 \tpm 0.56\\
& SNGP$_\text{I}$   & \textbf{3.96 \tpm 0.04} & 2.04 \tpm 0.02 & 9.88  \tpm 0.12 & 2.95  \tpm 0.02 & 10.54 \tpm 0.18 & 22.60 \tpm 0.30 & 94.94 \tpm 0.58\\
  \rule{0pt}{2.2ex}
10 & Base   & 4.70 \tpm 0.10 & 2.25 \tpm 0.02 &11.43 \tpm 0.16 & 3.42  \tpm 0.06& 17.17 \tpm 0.28 & 16.91 \tpm 0.38 &  93.63 \tpm 0.52\\
& Transfer & 4.49 \tpm 0.02& \textbf{2.07 \tpm 0.02} & 10.76 \tpm 0.05& 3.21 \tpm 0.01 & \textbf{12.40 \tpm 0.07} & 18.46 \tpm 0.27 &  \textbf{97.48 \tpm 0.55} \\
& SNGP$_\text{U}$   & 4.26 \tpm 0.03& 2.23 \tpm 0.03& 10.49 \tpm 0.22& 3.19 \tpm 0.02& 14.99 \tpm 0.13 & 20.19 \tpm 0.29 & 94.22 \tpm 0.89\\
&SNGP$_\text{I}$   & \textbf{4.23 \tpm 0.08} & 2.22 \tpm 0.05 & \textbf{10.47 \tpm 0.20} & \textbf{3.15 \tpm 0.02} & 13.85 \tpm 0.38 & \textbf{20.90 \tpm 0.16} & 94.35 \tpm0.65 \\
  \rule{0pt}{2.2ex}
5 & Base   & 5.04 \tpm 0.09& 2.41 \tpm 0.06& 12.33 \tpm 0.23& 3.67 \tpm 0.02& 23.73 \tpm 0.87  & 15.05 \tpm 0.80 & 90.79 \tpm 1.79\\
& Transfer & 4.94 \tpm 0.01& 2.25 \tpm 0.01 & 11.49 \tpm 0.02& 3.50 \tpm 0.01& 16.80 \tpm 0.03 & 15.86 \tpm 0.16 &  \textbf{97.12 \tpm 0.38}\\
& SNGP$_\text{U}$   & 4.43 \tpm 0.04 & 2.31 \tpm 0.02& 11.06 \tpm 0.08 & 3.36 \tpm 0.02& 18.92 \tpm 0.29 & 19.32 \tpm 0.29 & 91.60 \tpm 0.82\\\
& SNGP$_\text{I}$   & \textbf{4.41 \tpm 0.01} & \textbf{2.24 \tpm 0.02} & \textbf{10.92 \tpm 0.11} & \textbf{3.28 \tpm 0.01} & \textbf{16.30 \tpm 0.30} & \textbf{19.36 \tpm 0.48} & 93.17 \tpm 1.20\\
  \rule{0pt}{2.2ex}
3 & Base   & 5.41 \tpm 0.09& 2.48 \tpm 0.11& 12.99 \tpm 0.47& 3.88 \tpm 0.03& 28.85 \tpm 1.35 & 13.55 \tpm 0.47 & 90.96 \tpm 0.50 \\
& Transfer & 5.44 \tpm 0.01& 2.44 \tpm 0.07& 12.35 \tpm 0.04& 3.73 \tpm 0.01& 20.89 \tpm 0.04 & 13.81 \tpm 0.07 & \textbf{97.16 \tpm 0.33} \\
& SNGP$_\text{U}$   & 4.54 \tpm 0.04 &  2.34 \tpm 0.02& 11.31  \tpm 0.13 &3.50  \tpm 0.01 & 21.96 \tpm 0.34 & 17.78 \tpm 0.22 & 91.28 \tpm 0.82\\
& SNGP$_\text{I}$   & \textbf{4.51 \tpm 0.05} & \textbf{2.33 \tpm 0.04} & \textbf{11.06 \tpm 0.14} & \textbf{3.41 \tpm 0.01} & \textbf{18.04 \tpm 0.33} & \textbf{17.94 \tpm 0.20} & 92.49 \tpm 0.79\\
1 & Base   & 5.96 \tpm 0.26& 2.75 \tpm 0.04& 14.15 \tpm 0.62 & 4.46 \tpm 0.01 & 50.60 \tpm 0.99 & 11.33 \tpm 0.96 & 87.43 \tpm 3.49 \\
& Transfer & 6.52 \tpm 0.03& 2.95 \tpm 0.01& 14.30 \tpm 0.06 & 4.28 \tpm 0.01& 33.31 \tpm 0.12 & 10.02 \tpm 0.02 & \textbf{98.70 \tpm 0.03}\\
& SNGP$_\text{U}$   & 5.02 \tpm 0.05 &  2.53 \tpm 0.02& 12.26  \tpm 0.13 &3.96  \tpm 0.01 & 40.58 \tpm 0.50 & \textbf{15.14 \tpm 0.50} & 89.11 \tpm 0.83\\
& SNGP$_\text{I}$   & \textbf{5.00 \tpm 0.09} & \textbf{2.50 \tpm 0.02} & \textbf{12.14 \tpm 0.19} &\textbf{3.75 \tpm 0.02} & \textbf{26.63 \tpm 0.90} & 15.12 \tpm 0.39 & 90.34 \tpm 0.64\\
\bottomrule
\end{tabular}
\end{center}
\end{minipage}
\begin{minipage}[c]{\linewidth}
\centering
\includegraphics[width=0.9\linewidth]{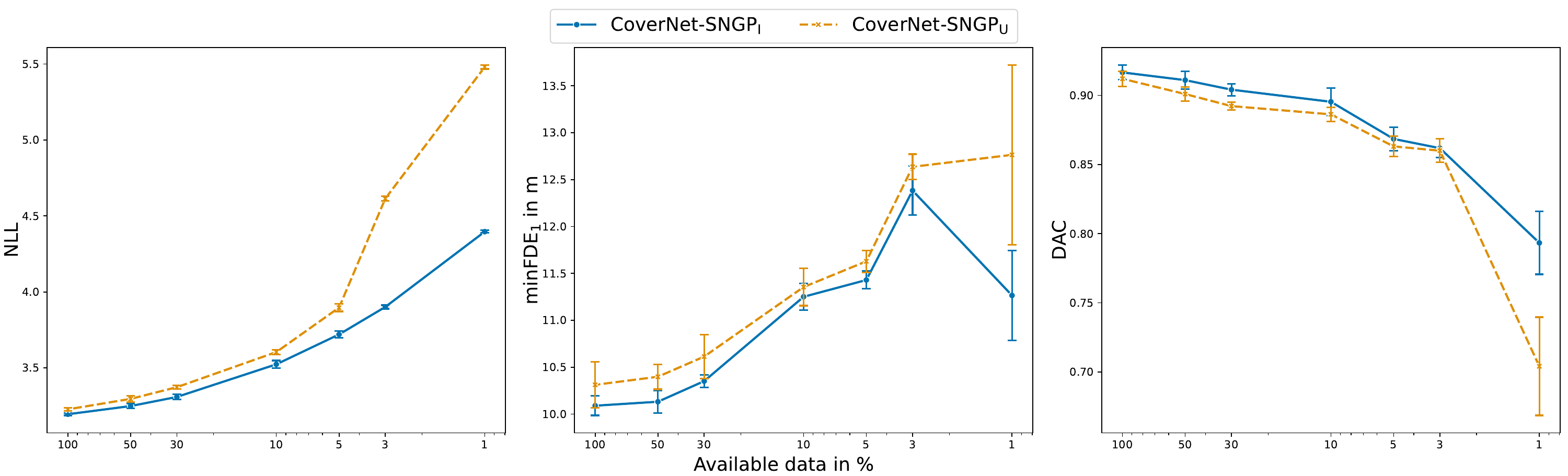}
\vspace{-5pt}
\caption{Average performance and standard deviation in NLL, \fde[n]{1} and DAC of five repetitions for the informed and non-informed CoverNet-SNGP over decreasing subsamples of NuScenes.}
\label{fig:performance_covernet_data}
\end{minipage}

\end{figure*}

Tab.~\ref{tab:data_nuscenes} and Tab.~\ref{tab:data_argo} show the performance of our CoverNet-SNGP$_\text{I}$ in comparison to the baselines on NuScenes and Argoverse2, respectively. We observe, that the prior drivability knowledge leads to notable performance benefits in our CoverNet-SNGP$_\text{I}$ and informed baselines (Transfer-CoverNet, GVCl-Det-CoverNet) across most metrics. The benefits from the prior drivability knowledge are most substantial in the calibration-sensitive metrics (RNK and notably NLL, e.g., as seen in Fig~\ref{fig:performance_covernet_data}) that directly benefit from the optimization in the knowledge tasks. The drivability knowledge is less helpful in discerning the best candidate between the remaining drivable candidate trajectories, leading to lower benefits in the respective metrics (\ade[n]{1}, \fde[n]{1}, ACC).

We also observe, that Transfer-CoverNet's benefits are limited to higher data regimes. In low data regimes, Transfer-CoverNet can even perform substantially worse than Base-CoverNet across all metrics (except DAC). In these low data regimes, Transfer-CoverNet may converge to less adequate minima, due to its weight initialization being overly biased towards drivability (illustrated by the rising DAC). In contrast, GVCL-Det-CoverNet and our CoverNet-SNGP$_\text{I}$ never decrease performance, with consistent benefits especially in low data regimes. This highlights a principal advantage of the \pil approach, where the informative prior helps to shape the complete loss landscape during training.

In comparison to GVCL-Det-CoverNet, our CoverNet-SNGP$_\text{I}$ shows benefits across most metrics, especially in low data regimes, even though both are trained using the \pil approach. The advantage is most visible in the metrics concerning the most-likely trajectory (\ade[n]{1}, ACC). CoverNet-SNGP$_\text{I}$ also shows more stable results with a lower standard deviations. Here, our CoverNet-SNGP$_\text{I}$ profits from using the full information of the posterior distribution at inference. 

\subsection{Effect of Location-Specific Training}
\label{sec:res_locs}

\begin{figure*}[t]
\tiny
\centering
\begin{minipage}[c]{\linewidth}
\captionof{table}{Average performance and standard deviation of 5 independent repetitions trained on Singapore and Boston locations from NuScenes.}
\vspace{-10pt}
\label{tab:locs_nuscenes}
\begin{center}
\begin{tabular}{lll*{7}{B}}

\textbf{Train Location} & \textbf{Model}  & \textbf{Test Location} & \ade[b]{1} & \ade[b]{5}& \fde[b]{1}  & \textbf{NLL} & \textbf{RNK} & \textbf{ACC} & 
 \textbf{DAC} \\
\midrule
Singapore & Base  &  Singapore& 5.33 \tpm 0.40 & 2.37 \tpm0.06 & 11.54\tpm 0.80  & 3.69\tpm 0.06 & 19.79\tpm 0.71 & 12.97\tpm 1.89 &  84.94\tpm 1.35\\
&  &Boston& 5.83 \tpm0.22 & \textbf{2.64 \tpm 0.04} & 12.76\tpm 0.41 & 3.93\tpm 0.07 & 24.98\tpm 1.08 & 10.18\tpm 0.80 & 89.79\tpm 2.13\\
& Transfer& Singapore & 5.47 \tpm 0.07 & 2.35 \tpm0.03 & 11.41\tpm 0.14  & 3.49\tpm 0.02 & 13.79\tpm 0.18& 11.49\tpm 0.50 & \textbf{94.29\tpm 0.59} \\
& &Boston& 6.65  \tpm 0.10 & 2.94 \tpm 0.03 & 14.26 \tpm 0.20 & 4.09 \tpm 0.01 & 24.09 \tpm 0.31 & 8.55 \tpm 0.40 & 
\textbf{96.09\tpm 0.20} \\
& SNGP$_\text{U}$ &  Singapore& 4.48 \tpm 0.06& 2.26 \tpm 0.02 & 10.05 \tpm 0.16 & 3.38 \tpm 0.03 & 15.06 \tpm 0.31  & \textbf{15.85 \tpm 0.46}& 85.30 \tpm 1.01 \\
&&Boston& 5.38 \tpm0.15 & 2.71 \tpm0.05 & 12.20 \tpm 0.35 & \textbf{3.65 \tpm 0.02} & \textbf{20.81 \tpm 0.56} & \textbf{13.15\tpm 0.73} & 90.37\tpm 0.95\\
& SNGP$_\text{I}$  & Singapore& \textbf{4.43 \tpm 0.07} & \textbf{2.20 \tpm 0.06}  & \textbf{9.83 \tpm 0.15} & \textbf{3.31 \tpm 0.06}  & \textbf{13.64 \tpm 1.31}& 15.84\tpm 1.05 & 86.56\tpm 0.50 \\
&&Boston& \textbf{5.36 \tpm 0.09} & 2.68 \tpm 0.08 & \textbf{12.18 \tpm 0.26} & \textbf{3.65 \tpm 0.01} & 21.56\tpm 1.40 & 12.95 \tpm 0.74 & 90.68 \tpm 0.79\\
  \rule{0pt}{2.5ex}
Boston  & Base & Boston  &  5.02 \tpm 0.20 & 2.32 \tpm 0.09 & 11.18\tpm 0.49 & 3.57 \tpm 0.08 & 18.10\tpm 1.31 & 13.18\tpm 1.23 &  90.18 \tpm 2.13 \\
&  &Singapore & 5.69 \tpm0.28 & 2.73 \tpm0.15 & 12.77 \tpm0.78 & 3.88 \tpm 0.07 & 23.42 \tpm 0.47 & 11.37 \tpm 0.98 & 82.03 \tpm2.92 \\
&Transfer & Boston & 4.78 \tpm0.06 & 2.19 \tpm0.01 & 10.21 \tpm0.12 & 3.41 \tpm 0.01 & 14.39 \tpm 0.19 & 14.02 \tpm 0.56  & \textbf{96.50 \tpm0.74}  \\
&&Singapore& 5.63 \tpm 0.06 & 2.64 \tpm 0.04 & 12.17 \tpm 0.20 & 3.70 \tpm 0.01 & 18.77 \tpm 0.16 & 11.40 \tpm 0.61 & \textbf{93.10 \tpm1.15} \\
& SNGP$_\text{U}$  &Boston & 4.62 \tpm 0.10 & 2.23 \tpm 0.02 & 10.46 \tpm 0.25& 3.32 \tpm 0.01 & 14.83 \tpm 0.40 & 16.57 \tpm 0.64 & 93.31 \tpm 0.40  \\
&&Singapore& 4.94 \tpm 0.07 & 2.61 \tpm 0.11 & 11.27 \tpm 0.17 & 3.58 \tpm 0.03 & 19.48 \tpm 0.16 & 14.93 \tpm 1.06 & 83.28 \tpm 0.60 \\
&SNGP$_\text{I}$& Boston   & \textbf{4.50 \tpm 0.04} &  \textbf{2.19 \tpm 0.02} & \textbf{10.13 \tpm 0.11}& \textbf{3.26 \tpm 0.02} & \textbf{12.97 \tpm 0.33} & \textbf{16.94 \tpm 0.60}& 94.01 \tpm 0.27 \\
&&Singapore& \textbf{4.82 \tpm 0.07} & \textbf{2.60 \tpm 0.07} & \textbf{10.95 \tpm 0.18} & \textbf{3.52 \tpm 0.04} & \textbf{18.39 \tpm 0.66} & \textbf{15.60 \tpm 0.70} & 85.36 \tpm 1.18 \\
\bottomrule
\end{tabular}
\end{center}
\vspace{0.5mm}
\end{minipage}

\begin{minipage}[c]{\linewidth}
\captionof{table}{Average performance and standard deviation of 5 independent repetitions trained on Palo-Alto and Miami locations from Argoverse2.}
\vspace{-10pt}
\label{tab:locs_argo}
\begin{center}
\begin{tabular}{lll*{7}{B}}

\textbf{Train Location} & \textbf{Model}  & \textbf{Test Location} & \ade[b]{1} & \ade[b]{5}& \fde[b]{1}  & \textbf{NLL} & \textbf{RNK} & \textbf{ACC} & 
 \textbf{DAC} \\
\midrule
Palo-Alto & Base  &  Palo-Alto& 4.94 \tpm 0.12 & 2.35 \tpm0.05 & 12.13 \tpm 0.20  & 3.45 \tpm 0.07 & 17.41 \tpm 0.21 & 14.72 \tpm 1.01 &  92.94 \tpm 1.41\\
&  &Ex-Palo-Alto&  5.02 \tpm 0.42 & 2.51 \tpm 0.23 & 12.24 \tpm 0.51 &  3.65 \tpm 0.12&  22.18 \tpm 1.20&  14.18 \tpm 0.79 & 91.90 \tpm 1.53 \\
& Transfer& Palo-Alto & 4.91 \tpm 0.05 & \textbf{2.19 \tpm0.01} & 11.32 \tpm 0.13  & 3.27 \tpm 0.01 & 13.75 \tpm 0.13& 18.66 \tpm 0.43 & \textbf{95.92 \tpm 0.38} \\
& &Ex-Palo-Alto&  5.33 \tpm 0.03&  2.44 \tpm 0.01& 12.39 \tpm 0.90&  3.63 \tpm 0.01&  18.30 \tpm 0.13& 13.68 \tpm 0.34 & \textbf{95.92 \tpm 0.46} \\
& SNGP$_\text{U}$ &  Palo-Alto& \textbf{4.23 \tpm 0.06}& 2.20 \tpm 0.01 & 10.63 \tpm 0.19 & 3.11 \tpm 0.03 & 15.03 \tpm 0.56  & \textbf{23.42 \tpm 0.35}& 92.02 \tpm 1.74 \\
&&Ex-Palo-Alto&  \textbf{4.55 \tpm 0.05}&  2.38 \tpm 0.02& 11.35 \tpm 0.13&  3.37 \tpm 0.02&  18.66 \tpm 0.55&  \textbf{18.58 \tpm 0.68} & 92.06 \tpm 1.55\\
& SNGP$_\text{I}$  & Palo-Alto& \textbf{4.23 \tpm 0.05} & \textbf{2.19 \tpm 0.04}  & \textbf{10.40 \tpm 0.22} & \textbf{3.06 \tpm 0.03}  & \textbf{13.72 \tpm 0.54}& 22.38 \tpm 0.47 & 91.74 \tpm 3.01 \\
&&Ex-Palo-Alto&  4.57 \tpm 0.11 & \textbf{2.37 \tpm 0.04} & \textbf{11.30 \tpm 0.32}&  \textbf{3.35 \tpm 0.01}&  \textbf{17.43 \tpm 0.54}&  18.09 \tpm 0.74 & 91.88 \tpm 2.25\\
  \rule{0pt}{2.5ex}

Miami  & Base & Miami &  4.02 \tpm 0.21 & 2.22 \tpm 0.11 & 10.28 \tpm 0.35 & 3.45 \tpm 0.06 & 14.12 \tpm 0.78 & 18.97 \tpm 0.31 &  95.20 \tpm 0.98\\
&  &Ex-Miami &  4.29 \tpm 0.22 & 2.31 \tpm 0.13 & 11.01 \tpm 0.39 &  3.47 \tpm 0.09&  16.18 \tpm 0.92&  17.92 \tpm 0.79 & 94.99 \tpm 1.12 \\
&Transfer & Miami & 3.91 \tpm0.01 & \textbf{1.85 \tpm0.01} & \textbf{9.52 \tpm0.02} & \textbf{2.94 \tpm 0.01} & \textbf{9.17 \tpm 0.04} & 21.33 \tpm 0.29  & \textbf{97.42 \tpm0.72}  \\
&&Ex-Miami & 4.31 \tpm 0.02 & \textbf{2.07 \tpm 0.01} & 10.47 \tpm 0.05&  3.10 \tpm 0.01&  \textbf{10.62 \tpm 0.04}& 19.65 \tpm 0.35  & \textbf{97.41 \tpm0.98} \\
& SNGP$_\text{U}$  &Miami & \textbf{3.88 \tpm 0.04} & 2.03 \tpm 0.02 & 9.74 \tpm 0.11& 3.00 \tpm 0.01 & 11.48 \tpm 0.16 & \textbf{22.07 \tpm 0.41} & 95.58 \tpm 0.40 \\
&&Ex-Miami &  \textbf{4.15 \tpm 0.04} &  2.21 \tpm 0.02& 10.44 \tpm 0.13&  3.11 \tpm 0.01&  13.56 \tpm 0.21& \textbf{21.50 \tpm 0.51} & 94.81 \tpm0.35 \\
&SNGP$_\text{I}$& Miami   & \textbf{3.88 \tpm 0.05} &  1.99 \tpm 0.02 & 9.65 \tpm 0.15& 2.99 \tpm 0.01 & 10.75 \tpm 0.21 & 21.71 \tpm 0.53 & 95.21 \tpm 0.46 \\
&&Ex-Miami &  4.17 \tpm0.05& 2.20 \tpm0.03 & \textbf{10.42 \tpm 0.15} & \textbf{3.09 \tpm 0.02} & 12.68 \tpm 0.31  &  21.25 \tpm 0.59& 94.26 \tpm 0.58 \\
\bottomrule
\end{tabular}
\end{center}
\vspace{0.5mm}
\end{minipage}

\begin{minipage}[c]{\linewidth}
\includegraphics[width=\linewidth]{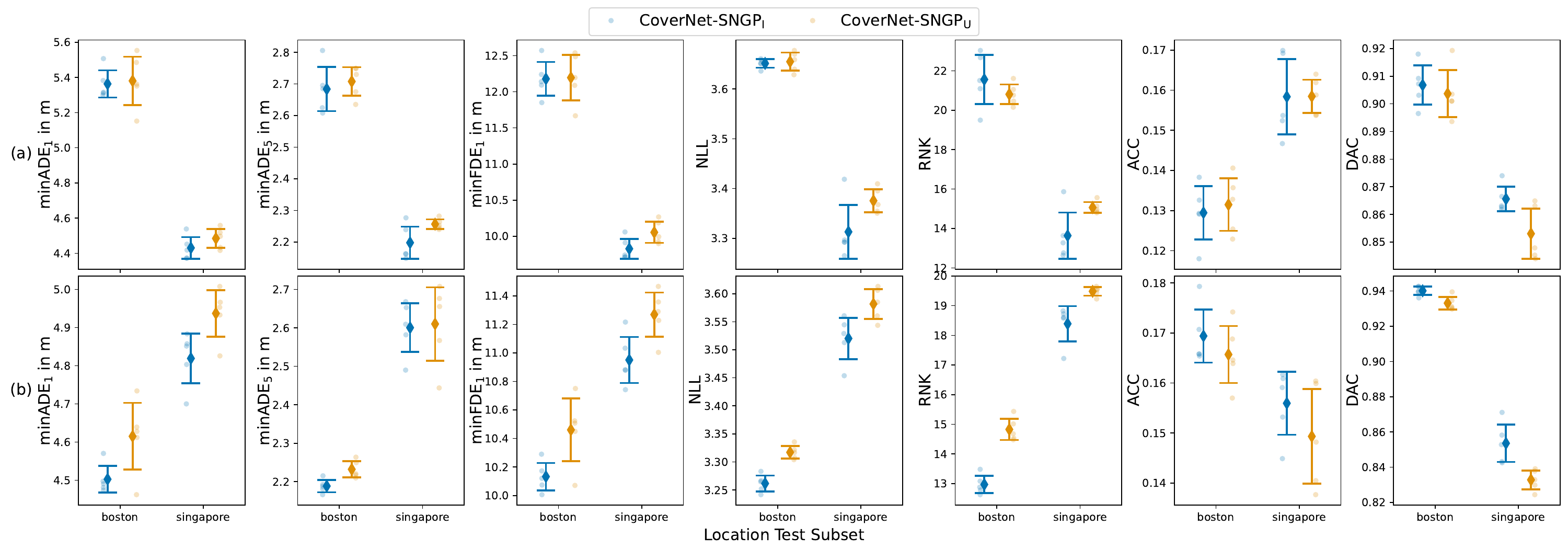}
\caption{Average performance and standard deviation of the informed and non-informed CoverNet-SNGP on Boston and Singapore test data, with (a) models trained on Singapore training data and (b) models trained on Boston training data (five repetitions).}
\label{fig:performance_covernet_loc}
\end{minipage}
\end{figure*}

Tab.~\ref{tab:locs_nuscenes} and Tab.~\ref{tab:locs_argo} show location-specific performances of our CoverNet-SNGP$_\text{I}$ in comparison to the baselines on NuScenes and Argoverse2, respectively. We observe, that the performance generally and substantially deteriorates in locations which are not included in the training data. This sensitivity of trajectory prediction models to location-transfers can be a major limitation to their practical use.

We also observe, that our CoverNet-SNGP$_\text{I}$ can help to alleviate this issue by consistently improving the generalization over location-transfers. This is most visible in the comparison of the Boston trained models on NuScenes (see Fig.~\ref{fig:performance_covernet_loc}) and the Palo-Alto trained models in Argoverse2, where we see a better performance across most metrics in same-location and location-transfer tests. The Transfer-CoverNet baseline performs even worse than Base-CoverNet in these cases, pointing to the same limitation we see in Sec.~\ref{sec:res_data} regarding its bias. In the other two comparisons, CoverNet-SNGP$_\text{I}$ still shows advantages (notably NLL). However, in case of Miami in Argoverse2, more training data is available (compare Sec.~\ref{sec:res_data}), and in case of Singapore in NuScenes the drivability knowledge might be less useful (see Fig.~\ref{fig:performance_covernet_loc}), since all models achieve a lower DAC.

\section{Conclusion}

Our work introduces a novel regularization-based continual learning method for the SNGP model. We apply this method in a \pil approach for trajectory prediction in autonomous driving, deriving a compute-efficient informed CoverNet-SNGP model integrating prior drivability knowledge. We demonstrate on two public datasets, that our informed CoverNet-SNGP increases data-efficiency and robustness to location-transfers, outperforming informed and non-informed baselines in low data regimes. Thus, we show that our proposed continual learning method is a feasible way to regularize SNGPs using informative priors. In future work, we plan to apply informed SNGPs to more recent transformer-based prediction models using self-supervised learning and investigate robustness against adversarial attacks and outliers. 

\appendix



\section*{Acknowledgments}

The research leading to these results is funded by the German Federal Ministry for Economic Affairs and Climate Action within the project ``KI Wissen -- Entwicklung von Methoden für die Einbindung von Wissen in maschinelles Lernen''. The authors would like to thank the consortium for the successful cooperation.

\bibliographystyle{named}
\bibliography{biblio}

\end{document}